\documentclass[runningheads]{llncs}

\usepackage[mobile]{eccv}

\usepackage{eccvabbrv}
\usepackage{graphicx}
\usepackage[accsupp]{axessibility}  
\usepackage{capt-of,color,colortbl,xspace,enumitem}
\usepackage{amsmath,amssymb,amsbsy,amsfonts,dsfont,pifont,bm,bbm,mathrsfs,mathtools,nicefrac}
\usepackage{algorithm,algpseudocode,listings}
\usepackage{booktabs,multirow,adjustbox,diagbox,threeparttable,tabularray}

\usepackage[pagebackref,breaklinks,colorlinks,citecolor=eccvblue]{hyperref}
\usepackage{hyperref}  
\usepackage{orcidlink}
\makeatletter
\DeclareRobustCommand\onedot{\futurelet\@let@token\@onedot}
\def\@onedot{\ifx\@let@token.\else.\null\fi\xspace}

\usepackage{orcidlink}
\usepackage{wrapfig}
\usepackage[capitalize]{cleveref}  
\crefname{section}{Sec.}{Secs.}
\Crefname{section}{Section}{Sections}
\crefname{table}{Tab.}{Tabs.}
\Crefname{table}{Table}{Tables}
\crefname{figure}{Fig.}{Figs.}
\Crefname{figure}{Figure}{Figures}
\crefname{equation}{Eq.}{Eqs.}
\Crefname{equation}{Equation}{Equations}
\newcommand{\tocite}[1]{{\color{red} [TO CITE]}}

\begin{document}

\title{CoReS: Orchestrating the Dance of \\
Reasoning and Segmentation}

\author{Xiaoyi Bao\inst{1,2,3,6} \and
Siyang Sun\inst{3} \and
Shuailei Ma\inst{4} \and
Kecheng Zheng\inst{5} \and
Yuxin Guo\inst{1,2,3} \and
Guosheng Zhao\inst{1,2,6} \and
Yun Zheng\inst{3} \and
Xingang Wang\inst{2,6}\thanks{Corresponding author.}}


\authorrunning{X.~Bao et al.}
\institute{
    \textsuperscript{\rm 1}School of Artificial Intelligence, University of Chinese Academy of Sciences\\
    \textsuperscript{\rm 2}Institute of Automation, Chinese Academy of Sciences\\
    \textsuperscript{\rm 3}Alibaba Group
    \textsuperscript{\rm 4}Northeastern University
    \textsuperscript{\rm 5}Ant Group\\
    \textsuperscript{\rm 6}Luoyang Institute for Robot and Intelligent Equipment\\
    \{baoxiaoyi2021,guoyuxin2021,zhaoguosheng2021,xingang.wang\}@ia.ac.cn,
    \{xiaomabufei,zkechengzk\}@gmail.com,
    \{siyang.ssy,zhengyun.zy\}@alibaba-inc.com
    \href{https://chain-of-reasoning-and-segmentation.github.io}{https://chain-of-reasoning-and-segmentation.github.io}
}
\maketitle

\begin{abstract}
The reasoning segmentation task, which demands a nuanced comprehension of intricate queries to accurately pinpoint object regions, is attracting increasing attention. However, Multi-modal Large Language Models (MLLM) often find it difficult to accurately localize the objects described in complex reasoning contexts. We believe that the act of reasoning segmentation should mirror the cognitive stages of human visual search, where each step is a progressive refinement of thought toward the final object. Thus we introduce the Chains of Reasoning and Segmenting (CoReS) and find this top-down visual hierarchy indeed enhances the visual search process. Specifically, we propose a dual-chain structure that generates multi-modal, chain-like outputs to aid the segmentation process. Furthermore, to steer the MLLM's outputs into this intended hierarchy, we incorporate in-context inputs as guidance. Extensive experiments demonstrate the superior performance of our CoReS, which surpasses the state-of-the-art method by 6.5\% on the ReasonSeg dataset.


\end{abstract}

\section{Introduction}\label{sec:intro}

Recently, multi-modal large language models (MLLM) have gained increasing recognition for their powerful capabilities in various tasks. By utilizing the knowledge repository contained in MLLM, humans can demand more complex multi-modal tasks than ever before. Among them, inference-based segmentation tasks can achieve more intelligent fine-grained understanding by combining traditional visual tasks with the reasoning process.

Currently, there are mainly two ways to handle this task: one is to equip MLLM with a segmentation decoder, and the other is to use LLM to output the mask in text form directly. For example, LISA~\cite{lai2023lisa} equips SAM~\cite{sam} decoder for LLaVA~\cite{liu2023visual} and constructs an inference segment database for training. VistaLLM~\cite{pramanick2023jack} directly uses LLM to generate text-formatted segmentation masks and designs an adaptive sampling algorithm to optimize the masks. 

\begin{figure}[tb]
  \centering
  \includegraphics[width=\linewidth]{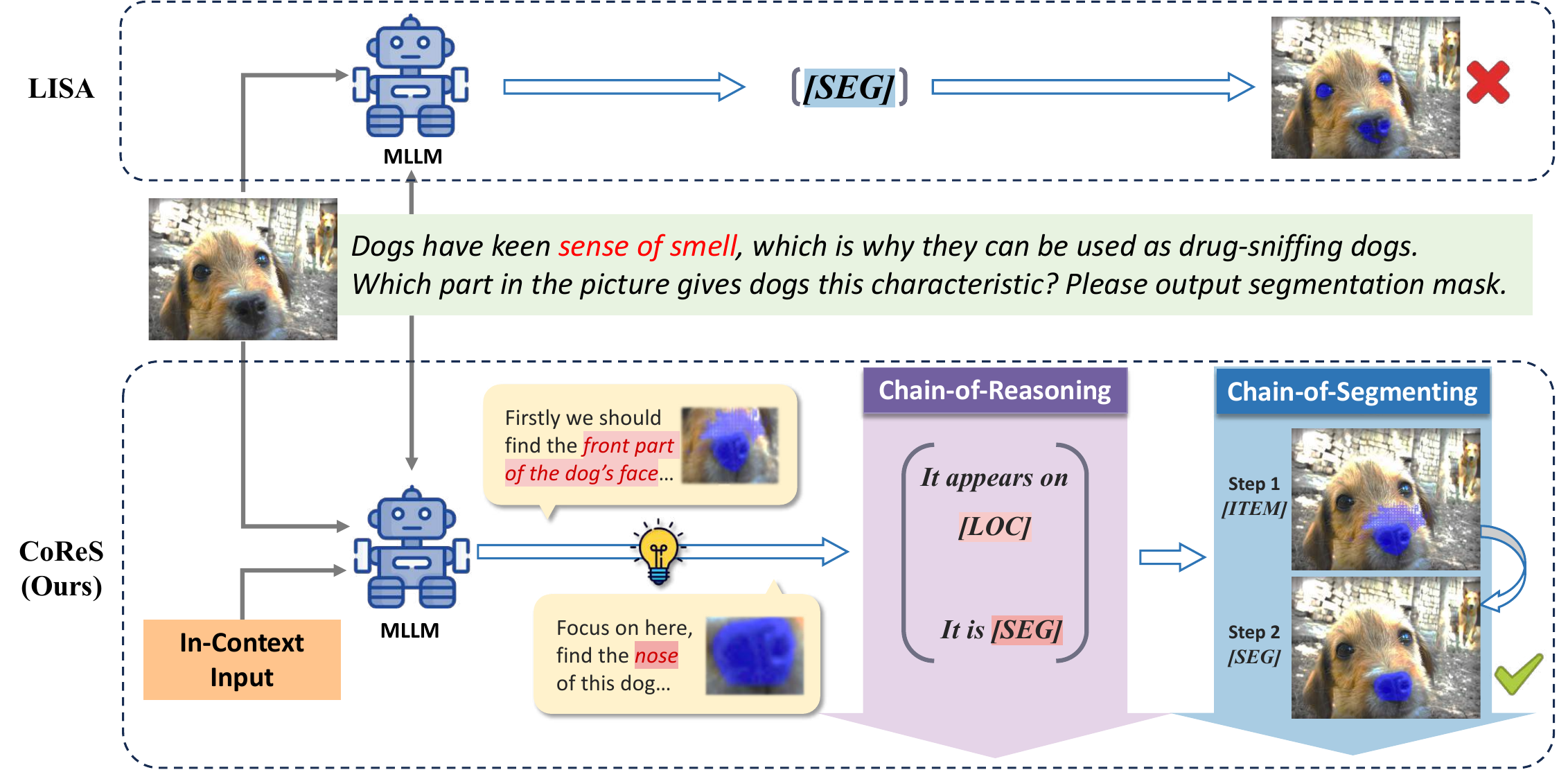}
  \caption{Comparison between our CoReS and LISA. \textbf{UP}: the process of LISA, \textbf{DOWN}: the diagram of CoReS. Given textual and visual inputs, LISA directly uses the [SEG] token output by MLLM to generate a mask. On the contrary, our CoReS involves breaking down the task of ``finding the part that gives dogs keen sense of smell'' into a logical chain such as ``first find the front part of the dog's face, then focus on this specific area, search for the nose of the dog.'' It can be observed that LISA incorrectly segments the dog's eyes, which are similarly round, dark, and important in sensory perception. In contrast, through in-context input and dual-chain structure, CoReS achieves the segmentation of the nose of the dog correctly. 
  }
  \label{motivation}
\end{figure}

The existing MLLM can effectively segment objects at the object level but struggles to differentiate objects referred to in reasoning text accurately. As shown in Fig.\ref{motivation}, when segmenting "part gives dogs sense of smell," the reasoning process directly searches for nose-like items that are round, dark, and important in sensory perception. Since the eyes of the dog have similar characteristics, LISA erroneously segments them as the object. The semantic similarity of such instances presents a substantial challenge to MLLM's ability to accurately localize and segment complex reasoning tasks.

How can we complete complex reasoning segmentation tasks? From the way humans handle similar tasks, we gain inspiration. The brain realizes object localization and searching typically in a top-down, targeted manner, where each step is a progressive refinement of thought toward the final object. Specifically, it is guided by pre-existing knowledge including a broad understanding of the typical positions of objects in the scene. Then a detailed analysis and synthesis of the specific item is followed to gradually approach the precise segmentation of objects. For example, when asked to find ``the wedding token usually exchanged by the groom and the bride'' in a picture, people draw upon their knowledge to presuppose that we should focus on the fingers where rings are usually worn. Therefore, people first identify the typical location of the hand, and then narrow the field of view to locate and segment the ring in the local area of the hand.

This top-down hierarchical structure can effectively help improve segmentation. Considering the disparities between modalities, how can we build such a visual logical hierarchy for multi-modal reasoning segmentation? Drawing inspiration from LLM's chain of thought, we propose the Chain of Reasoning and Segmentation (CoReS), a multi-modal chain of thought for fine-grained tasks, to establish such a unified visual hierarchy. As is shown in Fig.\ref{motivation}, it is a dual-chain structure aimed at decoupling the hierarchical thinking process. Specifically, the reasoning chain is reflected in the output of MLLM, injecting semantic information for different logical levels into different tokens in the chain. The segmentation chain utilizes the logic of the reasoning chain to iteratively optimize the segmentation results.

Since MLLM cannot actively extract outputs conforming to this top-down hierarchy without guidance, we propose adding extra in-context inputs for MLLM. This involves providing randomly sampled textual examples that indicate the desired chain-like rules of output. Although unrelated to the user's query input, these text-formed question-answer pairs contain the anticipated chain-like output rules. MLLM reads and transfers these rules to the output of multi-modal tasks, implicitly guiding the generation of the reasoning chain.

Our contributions can be summarized as follows:
\begin{itemize}
\item We propose CoReS, a multi-modal chain of thought. It provides a more accurate visual search for multi-modal fine-grained tasks through a top-down chain-like visual hierarchy.

\item To form a chain-like multi-modal process, we propose a dual-chain structure, integrating both modalities in the same visual hierarchy. Moreover, to lead MLLM to generate such a hierarchy actively, we provide guidance through in-context input. With these randomly sampled text-based question-answer pairs, MLLM learns the chain-like rule and transfers it to multi-modal tasks.

\item Extensive experiments demonstrate the effectiveness of our proposed method, which achieves state-of-the-art accuracy on ReasonSeg benchmarks.
\end{itemize}
\section{Related Work}\label{sec:related}
\subsection{Interactive Segmentation}
Traditional segmentation tasks use single images as input, requiring the segmentation of objects belonging to a set of predefined categories~\cite{fcn,deconvnet,segnet,unet,deeplab,dilation,parsenet,pspnet,icnet,denseaspp,danet,ccnet,psanet,asymmetric_nonlocal,cheng2021per,guo2024dual,mashuailei,lai2021semi,tian2022adaptive,tian2023learning}. These tasks are straightforward, with simple images, and distinct objects. Existing methods have already achieved impressive results in this area~\cite{kirillov2019panoptic,xiong2019upsnet,cheng2020panoptic,li2021fully}. To meet the demands of more complex task requirements, the field of segmentation research is gradually evolving from purely visual to multi-modal approaches.

One extension is the open-vocabulary segmentation task, which involves segmenting objects specified by class names that do not belong to a fixed set of categories~\cite{xu2023open}. Such tasks often employ visual language models like CLIP as a bridge between modalities~\cite{ding2022open,liang2023open}. To facilitate more convenient interaction, the referring segmentation task~\cite{kazemzadeh2014referitgame,nagaraja2016modeling} enables interaction with human language rather than the semantic category list, aiming to leverage explicit text description to segment the target object. 

To expand the diversity of interaction methods, existing works~\cite{kirillov2023segment, zou2023generalized, zou2023segment} design a variety of prompting methods including points, boxes, scribbles, noise masks, and so on. Kirillov.et.al \cite{kirillov2023segment} introduced SAM, trained with billions of high-quality masks, supporting bounding boxes and points as prompts while demonstrating exceptional segmentation quality. X-Decoder~\cite{zou2023generalized} and SEEM~\cite{zou2023segment} further support various human interaction methods. Recently, Lai. et.al proposes LISA~\cite{lai2023lisa}, simply integrated the MLLM with SAM \cite{kirillov2023segment} to tackle the reasoning segmentation task and enhance existing visual segments with self-reasoning abilities. Although LISA inherits the excellent abilities of large language models in text reasoning, its performance in segmentation, especially for hard-to-perceive part objects, is not satisfactory. 

\subsection{Multi-modal Large Language Model}

The multi-modal large language models integrate the LLM~\cite{gpt,touvron2023llama,chowdhery2023palm} and vision encoder~\cite{radford2021learning,sun2023eva} to transfer the reasoning ability and huge world knowledge for the vision tasks.
Flamingo~\cite{alayrac2022flamingo} proposes a cross-attention structure to integrate visual information into the nlp contexts, enabling visual in-context learning. 
Several works (such as BLIP-2~\cite{li2023blip}, mPLUG-OWL~\cite{ye2023mplug} Otter~\cite{li2023otter}, LLaVA~\cite{liu2023visual}, MiniGPT-4~\cite{zhu2023minigpt}, FROMAGe~\cite{koh2023grounding} and so on) leverage the visual encoder and projection module (such as linear projection, Qformer, Casual Qformer and so on) to translate the visual features and directly feed them into the LLM along with the NLP token embeddings.

Recently, the use of MLLMs is no longer limited to understanding tasks, and there has been an attempt to utilize them for fine-grained visual tasks, such as grounding and segmentation.
Kosmos-2~\cite{peng2023kosmos} constructs large-scale data of grounded image-text pairs, attempting to infuse grounding capabilities into LLMs. DetGPT~\cite{detgpt} bridges the fixed MLLM and open-vocabulary detector. GPT4RoI~\cite{zhang2023gpt4roi} introduces RoI visual features as input and trains the model on region-text pairs. LISA~\cite{lai2023lisa} integrates SAM~\cite{sam} as a segmentation decoder into MLLMs and proposes the reasoning segmentation task. VistaLLM~\cite{pramanick2023jack} proposes a unified framework for MLLMs with single and multiple visual scene inputs and introduces an adaptive sampling algorithm to refine the NLP format mask of the MLLM's output. In this paper, we design a novel chain-of-thought manner to enable MLLM to realize fine-grained reasoning tasks.

\subsection{Chain of Thoughts}

CoT, as a convenient and efficient method for enhancing complex reasoning in large language models (LLMs)~\cite{DBLP:journals/corr/abs-2201-11903}, has significantly improved their performance in generating rationales and inferring accurate answers in numerous domains, including commonsense and arithmetic. Existing CoT prompting for LLMs is primarily used in inference and can be categorized into two major paradigms: Zero-shot-CoT and Few-shot-CoT. Zero-shot-CoT~\cite{DBLP:journals/corr/abs-2205-11916} directly leverages a single prompt like ``Let’s think step by step'' to generate reasoning chains. Few-shot-CoT~\cite{DBLP:journals/corr/abs-2201-11903} uses reasoning demonstrations one by one as the prompt information. 

For MLLMs, due to limitations in model size and performance, directly employing these two kinds of CoT in inference does not lead to an effective enhancement. Therefore, multi-modal CoT~\cite{DBLP:journals/corr/abs-2302-00923} typically acquires this ability by fine-tuning the model on the constructed multi-modal CoT dataset. However, leveraging vision information effectively and fusing visual features with text representation in a multi-modal chain of thought (CoT) poses a significant challenge. Prior work~\cite{DBLP:journals/corr/abs-2209-09513} attempts to use image captions and incorporate them after text input, but this approach results in substantial information loss of images. To avoid the complexity of constructing extra multi-modal CoT dataset, KAM-CoT~\cite{kamcot} proposes a two-stage training process with knowledge graphs grounding to generate effective rationales and answers for reducing the computational cost and substantial hardware resources. Utilizing the concept of a thought chain, $V^*$~\cite{v++} combines LLM and MLLM to construct a visual search algorithm that surpasses the performance of GPT-4V. As far as we know, the multi-modal application of COT has been confined to visual understanding tasks like visual question answering and has not extended to dense prediction tasks.
\begin{figure}[tb]
  \centering
  \includegraphics[width=\linewidth]{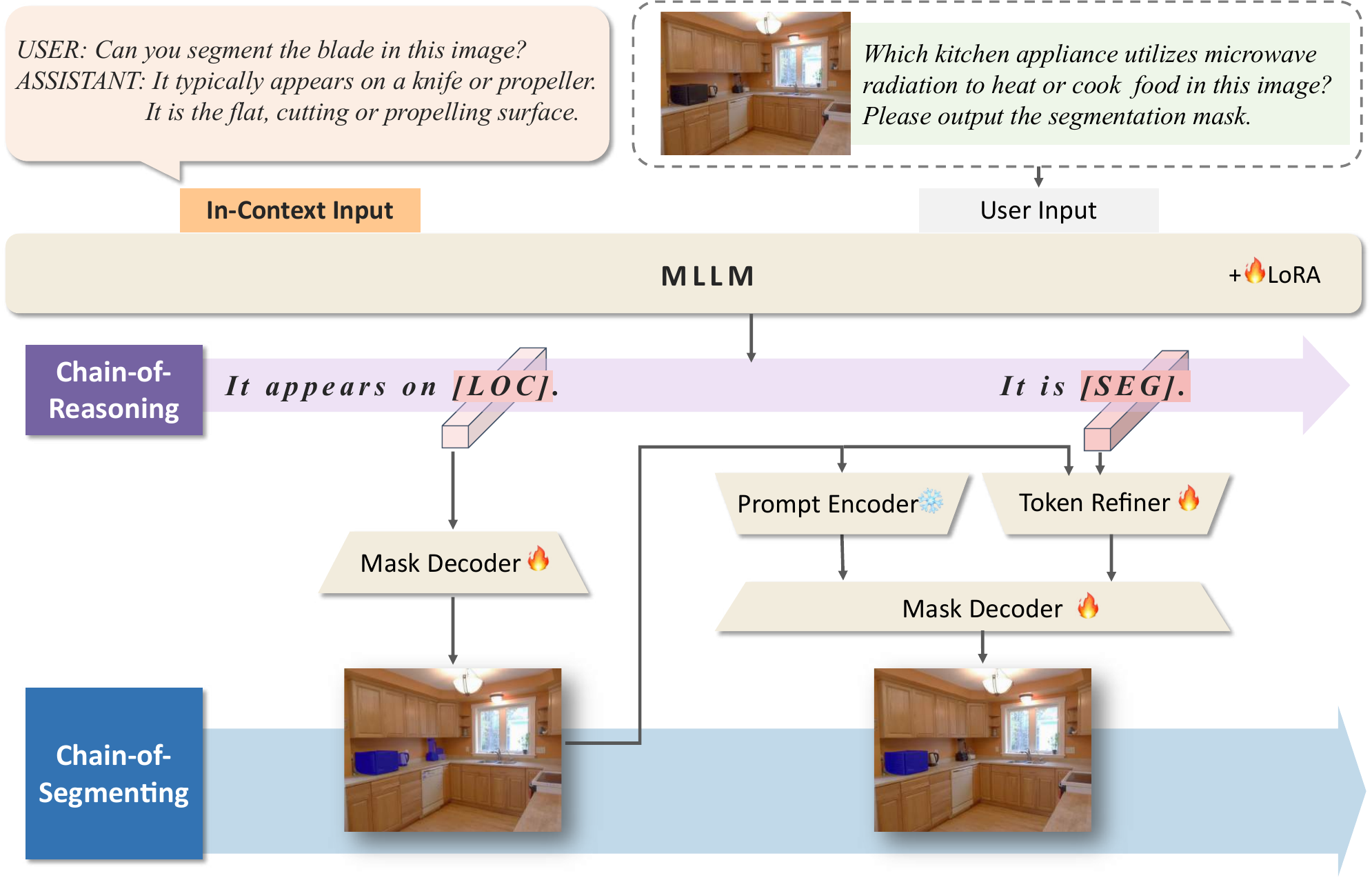}
  \caption{Overall architecture of CoReS. The input of MLLM consists of the user input in gray and the extra in-context input in orange, which consists of question-answer examples unrelated to the user query. MLLM generates output at the logical level of chain-of-reasoning, where the token embeddings of \texttt{[LOC]} and \texttt{[SEG]} serve as prompt inputs for different positions of the segmentation chain, guiding the chain to generate segmentation results progressively. We exclude the visual feature input to the mask decoder extracted by the extra vision backbone for conciseness here.
}
  \label{structure}
\end{figure}

\section{Method}\label{sec:method}
In line with the process of human visual search, a top-down visual hierarchy can aid the MLLM in progressively pinpointing objects referred to in reasoning texts. Inspired by this concept, we introduce CoReS, which applies the idea of the chain of thought from natural language processing to the execution of fine-grained tasks in a multi-modal context. As illustrated in Fig.\ref{structure}, the core of CoReS lies in its dual-modal, dual-chain structure, which consists of a chain-of-reasoning and a chain-of-segmenting. In addition to this, the extra in-context guidance plays a critical role in the formation of a chain-like hierarchy. Detailed explanations of these two components are provided in Sec.\ref{sec:dualchain} and Sec.\ref{sec:icl}.

\subsection{Dual-Modal Dual-Chain}\label{sec:dualchain}
\subsubsection{Chain-of-Reasoning:}
To achieve consistency between the modalities in the multi-modal chain of thought, the output of MLLM needs to conform to the top-down semantic logic required for fine-grained visual tasks. 

To this end, we use the kind of response templates like ``\texttt{It appears on [LOC]. It is [SEG].}'' By constraining the sentence structure of the MLLM outputs, implicit constraints are imposed on the logical chain output $H$ by MLLM. Given the image $Q_{img}$ and query input $Q_{text}$, this template of output compels MLLM to incorporate different information at different positions of the token, thus forming a semantic-level chain of reasoning. Specifically, MLLM injects the information about the scene or item that the object most likely to appear on into the \texttt{[LOC]} token while the information about the object itself is injected into the \texttt{[SEG]} token of the following sentence.

We fine-tune MLLM using LoRA and employ cross-entropy loss for the supervision of the chain of reasoning. The specific formula of ${L}_{CoR}$ is as follows.
\begin{equation}
            \mathcal{L}_{CoR} = \mathcal{L}_{CE}(\boldsymbol{p}(H|Q_{img}, Q_{text}),\boldsymbol{t}).
\end{equation}
$\boldsymbol{p}(H|Q_{img}, Q_{text})$ and $\boldsymbol{t}$ refer to output logits of MLLM and the token embedding of our chain-like template, respectively. $\mathcal{L}_{CE}$ is the cross-entropy loss.

\subsubsection{Chain-of-Segmenting:}
Upon obtaining text-form outputs that align with the visual hierarchy, the next task is to apply them as guidance for the visual modal. Thus we propose the Chain-of-Segmenting, iteratively generating segmentation results in a unified hierarchical manner. Obtaining mask annotations for intermediate logical layers is impractical due to the logic subjectivity, so MLLM with world knowledge is used for a unified unsupervised representation. The scene-level information injection of C1 is analyzed in the appendix.

During the hierarchical generation of segmentation, the segmentation result from the previous level $m^{t-1}$ serves as a cue. Since the \texttt{[LOC]} and $m^{t-1}$ in the first logical layer focus on common locations and do not entirely represent inclusion relationships, we choose its logical form to serve as a soft mask prompt, processed by the prompt encoder $\theta$ of the SAM framework.
\begin{equation}
            \mathcal{M}^{t} = \theta(m^{t-1}), t>=1.
\end{equation}
$\mathcal{M}^{t}$ is the output of prompt encoder from the $t$-th logical level. 

The implementation of text guidance follows the embedding-based approach in LISA. From the last layer feature output by MLLM, the embedding $(\boldsymbol{h}^0$, $\boldsymbol{h}^1 \in \mathbb{R}^{1\times 256})$ are extracted at positions of \texttt{[LOC]} and \texttt{[SEG]}. They serve as a prompt in the text modality to guide the corresponding segmentation process.

To further enhance the inter-modal connections in a top-down hierarchy, we propose to refine the token embedding of the reasoning chain with the help of the visual output $m^{t-1}$ from the previous logical level. Specifically, we use masked-average-pooling (MAP) to create a prototype. This prototype corresponds to the visual features activated at the current stage. We then use this prototype to calibrate text tokens at the next logical level. This process allows for iterative optimization of tokens generated in a single forward pass of the reasoning chain, aligned with the hierarchy.
\begin{equation}
            \hat{\boldsymbol{h}}^t = \mathcal{R}(\beta(\boldsymbol{h}^t),F_v(Q_{img}),m^{t-1}), t>=1.
\end{equation}
$\hat{\boldsymbol{h}}^t$ is the refined token embedding from the $t$-th logical level while $\boldsymbol{h}^t$ is the unrefined one in the chain-of-reasoning. $\beta$ is a multilayer perceptron (MLP) to project output features. We employ the SAM image encoder $F_v$ as the visual backbone. $\mathcal{R}$ refers to the token refiner whose working function can be expressed by the following formula.
\begin{equation}
            \mathcal{R}(h,i,m) = h+CA(h,MAP(i,m)).
\end{equation}
$CA$ and $MAP$ refer to cross-attention and masked average pooling, respectively.

Subsequently, the extracted features of the query image $Q_{img}$ are passed to the mask generator $\gamma$ as input. For brevity, the aforementioned process is omitted in Fig.\ref{structure}. In summary, the generation process of the segmentation chain can be expressed as follows.

\begin{equation}
            m^{t} = \gamma(F_v(Q_{img}),\mathcal{M}^t,\hat{\boldsymbol{h}}^t).
\end{equation}
where $m^{t}$ is the $t$-th level output in the segmenting chain. In the dual-level setting of the segmentation chain, $t=0/1$.

Parameters in the projection layer and mask decoder are set trainable, while the SAM image encoder and prompt encoder are kept frozen to keep the generalization ability. For the segmentation chain, we only provide supervision for the final mask generated at the last level of the chain. We choose dice loss and cross-entropy loss as the loss function, which can be represented as:
\begin{equation}
            \mathcal{L}_{CoS} = \lambda_d \mathcal{L}_{DICE}(m^T, M_{gt})+\lambda_c \mathcal{L}_{CE}(m^T, M_{gt}).
\end{equation}
where $m^T$ and $M_{gt}$ are the predicted final mask and the ground truth, respectively. In the dual-level setting of the segmentation chain, $T=1$. The analysis for the case when T > 1 can be found in Sec.4.4. The weights $\lambda_d$ and $\lambda_c$ are set to 0.5 and 2.0 empirically. 

As the following equation shows, the final loss is a weighted sum of the textual loss from the chain-of-reasoning $\mathcal{L}_{CoR}$ and the visual loss $\mathcal{L}_{CoS}$ from the chain-of-segmenting.
\begin{equation}
            \mathcal{L}_{total} =  \lambda_R\mathcal{L}_{CoR}+ \lambda_S\mathcal{L}_{CoS}.
\end{equation}
where the weights $\lambda_R$ and $\lambda_S$ are set to 0.5 empirically. The impact of these weights can be found in the Appendix.

\subsection{In-Context Guidance}\label{sec:icl}
The described hierarchical dual-chain structure progressively identifies the objects referred to in the reasoning text. However, if only sentence templates are used as the supervisory signal, the MLLM cannot actively uncover and output hierarchical logical relationships. 

\begin{figure}[tb]
  \centering
  \includegraphics[width=\linewidth]{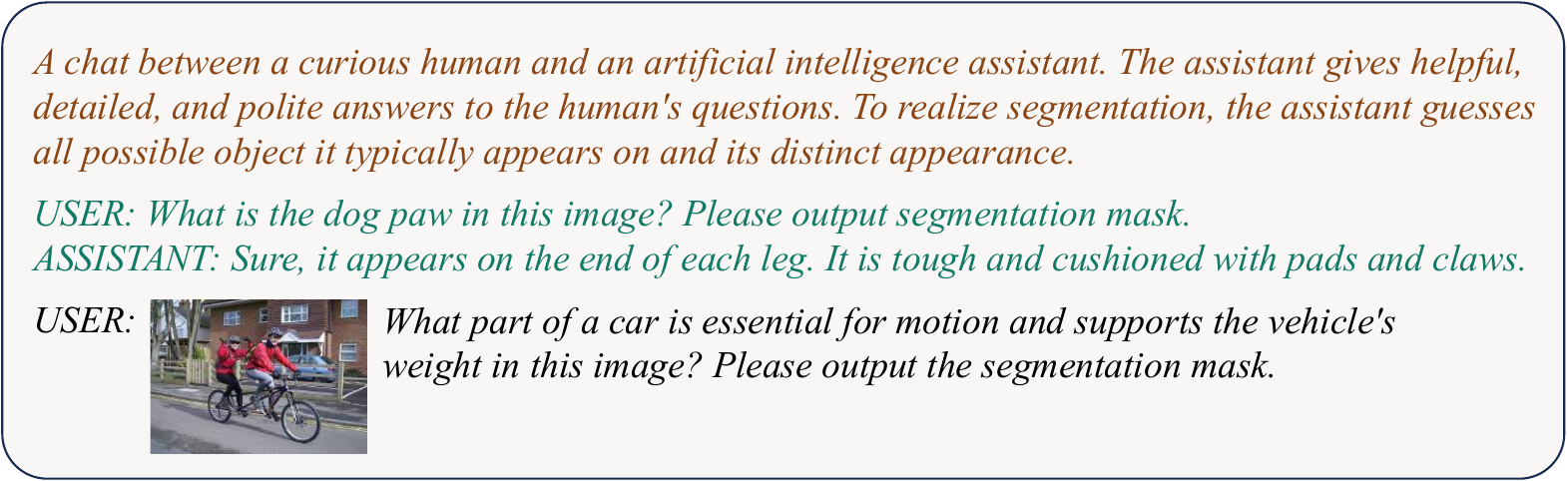}
  \caption{Examples of MLLM input. The brown section represents the system prompt commonly used for MLLM. The green section is the in-context input proposed in this paper, and the black section is the user input for questioning the image.
}
  \label{input}
\end{figure}

We thus propose a training paradigm involving in-context guidance to provide logical cues to the output of MLLM. 
To present the required top-down output logic rules to the MLLM, we introduce some contextually provided examples as guidance, before the initial user input of image-query pairs, as depicted in Fig.\ref{input}. 

Specifically, we pre-construct a context library in a pure text format. This process is to induce ChatGPT to generate more question-answer pairs with several manually written examples. For the question part of the examples, we utilize the same questioning format. For the sake of conciseness, CoReS does not include corresponding images and directly poses questions based on the name of the segmented object. For the answer format, no special tokens are used in the examples; instead, normal replies are given based on the logical aspects of where the questioning object usually appears and its common features.

During each forward pass in CoReS, we randomly draw examples from the context library to serve as in-context input. The reason we opt not to generate query-specific textual prompts lies in the high computational and parameter demands. The two-stage method requires forward passes through two different MLLMs, which is quite inefficient.

Examples extracted in this manner are unrelated to the current query image but implicitly embed the logic rules for output. As a rule provider, this in-context input serves as a prompt for MLLM to inject specific information into specific token positions. For instance, the token at the \texttt{[LOC]} position should be injected with location-related information such as ``a knife or propeller'', while the token at the \texttt{[SEG]} position should be injected with information regarding ``the flat, cutting or propelling surface'' of the object. In this way, MLLM extracts the top-down rules from pure textual context and transfers it to the output for new instances referred to by multi-modal queries.

Unlike in-context learning traditionally used solely to enhance the inference performance, our in-context question-answer pairs also play a role in the training process. Through this contextual prompting, CoReS teaches the model what semantic logic to follow in generating answers during the tuning phase.
\section{Experiments}\label{sec:exp}

\subsection{Implementation Details}
\textbf{For training datasets,} we follow LISA to utilize a combination of datasets for semantic, referring, and reasoning segmentation while abandoning the use of the visual-question-answering dataset. Related reasons and details can be found in the appendix. \textbf{For the test dataset}, we conduct experiments on the ReasonSeg, which comprises image-question pairs with reasoning difficulty and the ground truth segmentation masks. \textbf{For evaluation metrics,} we use graph-averaged Intersection-over-Union (gIoU), which is the most commonly used metric in referring segmentation. The cumulative Intersection-over-Union (cIoU) is used as an auxiliary indicator, despite its bias to large-area targets of images. We conduct experiments in the Appendix to clarify this bias.

The multi-modal LLM, unless specifically indicated otherwise, refers to the LLaVA-7B-v0. Moreover, we use SAM-ViT-H as the image backbone. Other specific implementation details are presented in the Appendix.

\begin{table}[!t]
\caption{Performance comparison between CoReS and other methods on ReasonSeg. ``ft'' denotes using the reasoning segmentation dataset to fine-tune the model.}
\label{mainexcel}
\centering
\setlength{\tabcolsep}{8pt}
\begin{threeparttable}
\begin{tabular}{l|c|c|c|c|c}
\toprule[1pt]
\multirow{2}{*}{Method}&Text&ReasonSeg&\multicolumn{3}{c}{ReasonSeg|test} \\
 &Model&val|overall&short&long&overall \\
\midrule
OVSeg~\cite{liang2023open}&CLIP ViT-L&28.5&18.0&28.7&26.1 \\
GRES~\cite{liu2023gres}&BERT&22.4&17.6&22.6&21.3 \\
X-Decoder~\cite{zou2023generalized}&UniCL&22.6&20.4&22.2&21.7\\
SEEM~\cite{zou2023segment}&UniCL&25.5&20.1&25.6&24.3\\
\midrule
LISA~\cite{lai2023lisa}&&44.4&37.6&36.6&36.8   \\
CoReS&&\cellcolor{gray!20}54.8&\cellcolor{gray!20}{41.0}&\cellcolor{gray!20}{50.9}&\cellcolor{gray!20}{48.7} \\
LISA(ft)&&52.9&40.6&49.4&47.3   \\
CoReS(ft)&\multirow{-4}{*}{LLaVA-7B}&\cellcolor{gray!20}\textbf{59.4}&\cellcolor{gray!20}\textbf{44.2}&\cellcolor{gray!20}\textbf{55.0}&\cellcolor{gray!20}\textbf{52.4} \\
\midrule
LISA(ft)&&56.2&44.3&54.0&51.7   \\
CoReS(ft)&\multirow{-2}{*}{LLaVA-13B}&\cellcolor{gray!20}\textbf{61.8}&\cellcolor{gray!20}\textbf{49.7}&\cellcolor{gray!20}\textbf{58.3}&\cellcolor{gray!20}\textbf{55.9} \\
\midrule
LISA(ft)&LLaVA-&65.0&55.4&63.2&61.3  \\
CoReS(ft)&-v1.5-13B&\cellcolor{gray!20}\textbf{68.1}&\cellcolor{gray!20}\textbf{57.9}&\cellcolor{gray!20}\textbf{66.4}&\cellcolor{gray!20}\textbf{65.5} \\
\bottomrule[1pt]     
\end{tabular}
\end{threeparttable}
\end{table}

\subsection{Comparison with State-of-the-Arts}
We compare the performance of CoReS with other methods on the ReasonSeg dataset, as shown in Tab.\ref{mainexcel}. Our approach improves by approximately 35\% compared to the multi-modal grounding models that do not use LLM. 
\begin{figure}[!t]
  \centering
  \includegraphics[width=0.9\linewidth]{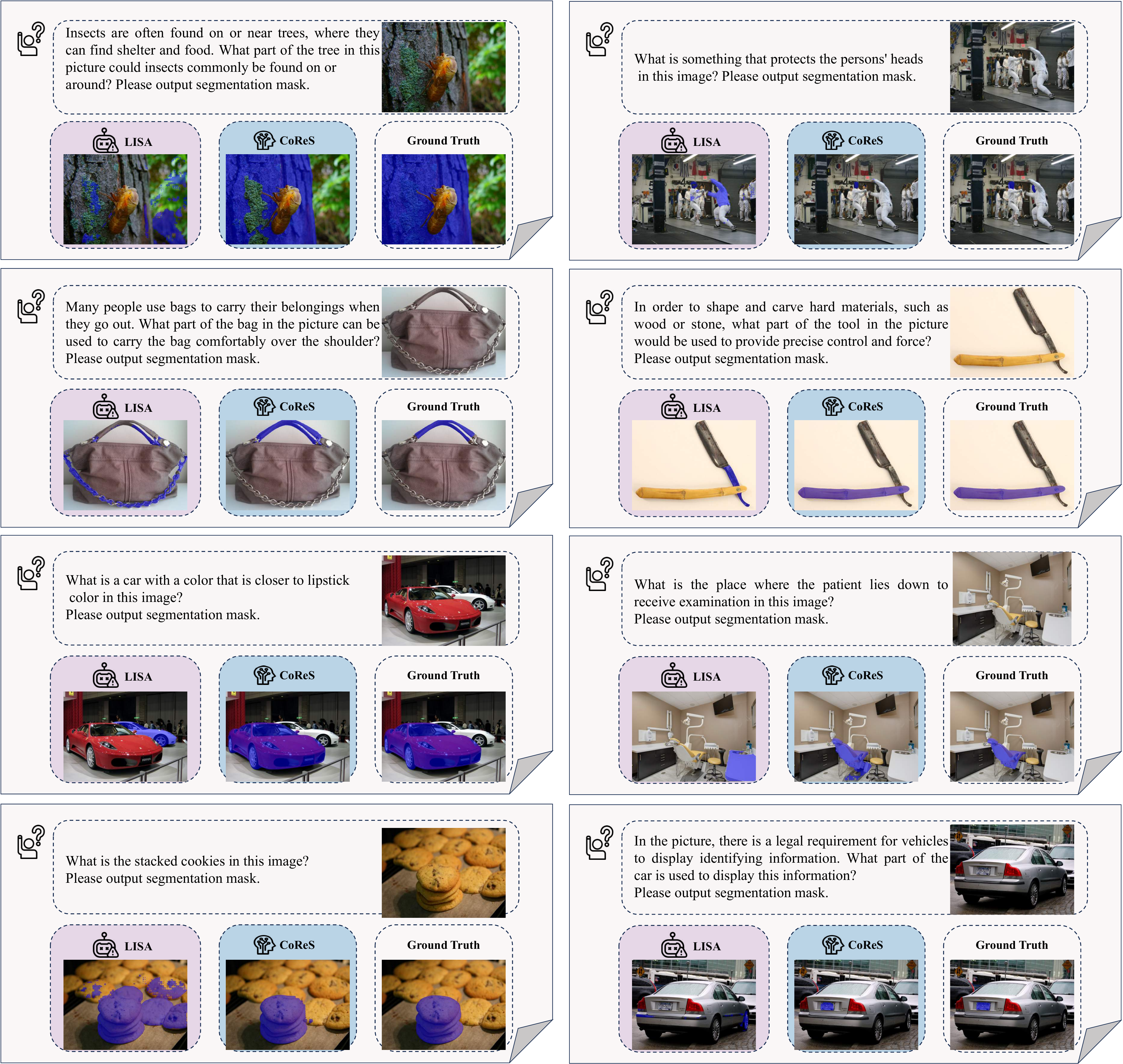}
  \caption{Visual comparison of CoReS and LISA. 
}
  \label{sample}
\end{figure}
Furthermore, we also compare the performance of LISA, a baseline method that similarly incorporates world knowledge by MLLM. This comparison with CoReS is made using MLLM of different scales. Without fine-tuning the ReasonSeg dataset, our method outperforms LISA by 10.4\%, and this margin remains 6.5\% after fine-tuning. It is worth noting that the un-tuned version of CoReS outperforms the fine-tuned version of LISA by 1.9\%. This demonstrates that the performance improvement brought by CoReS does not rely on provided training data but fundamentally leverages the potential of the MLLM more efficiently. When using LLaVA-13B and LLaVA-v1.5-13B as MLLMs, CoReS also outperforms LISA by a considerable margin. Achieving the same level of improvement becomes more challenging on a higher-performing base than LISA, thus the slight decrease in the magnitude of improvement is reasonable. Overall, these significant performance improvements prove the efficacy of CoReS. 

\begin{figure}[!t]
  \centering
  \includegraphics[width=0.9\linewidth]{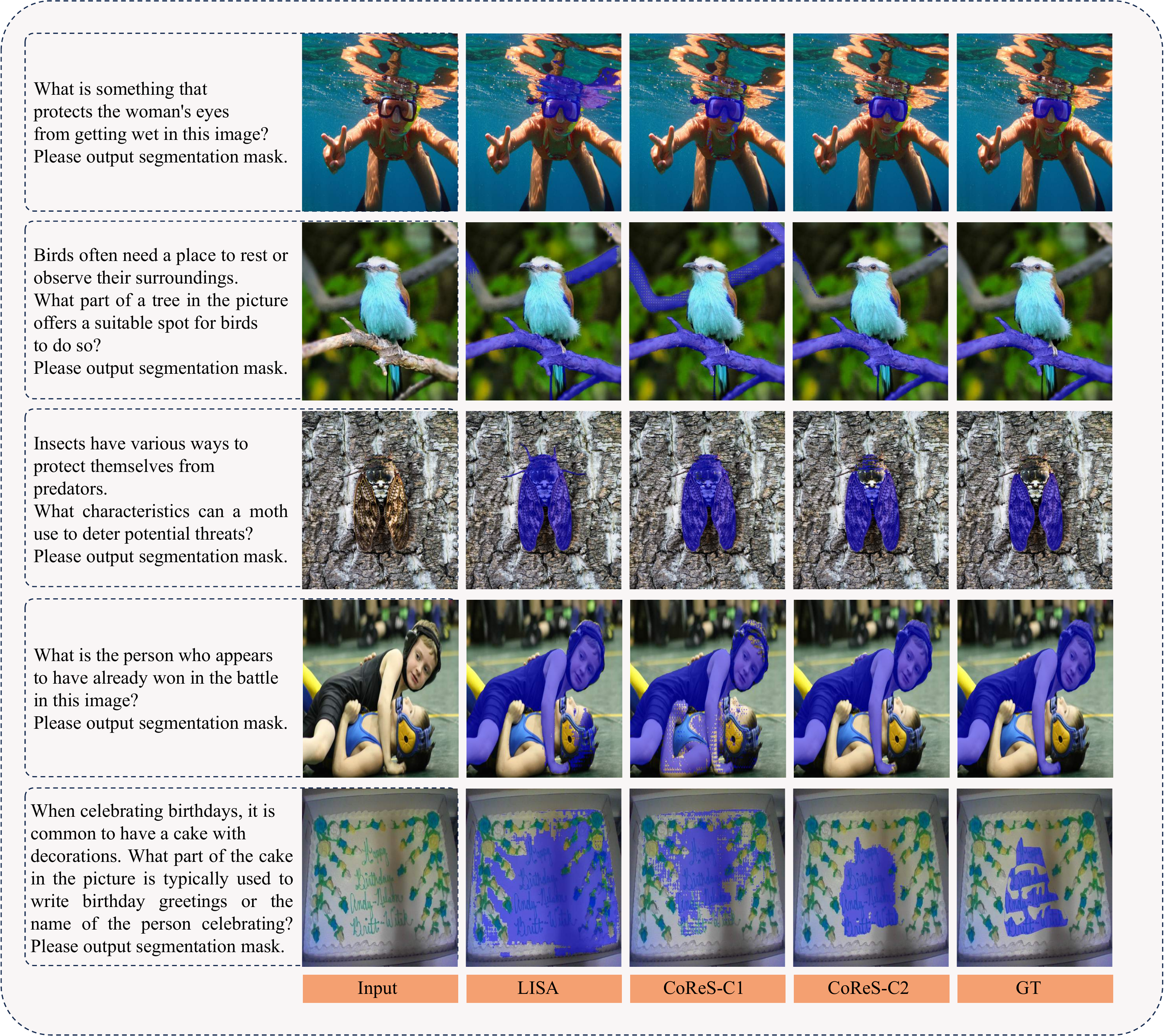}
  \caption{Qualitative interpretation of the advantages of the multi-modal chain-of-thought over LISA. From left to right are the input image, LISA result, CoReS first logic layer segmentation result, CoReS final result, and ground truth mask.}
  \label{5pic}
\end{figure}

Qualitative results also prove the effectiveness of CoReS in Fig.\ref{sample}, Fig.\ref{5pic}, and the appendix. As shown in Fig.\ref{sample}, when facing complex reasoning problems, LISA exhibits errors in grounding the exact instance referred by the reasoning query, while CoReS achieves the correct answers. Fig.\ref{5pic} provides a more detailed demonstration of the effects of the multi-modal chain of thought. Whether it is complex reasoning, such as ``moth evading predators,'' or segmentation difficulties caused by underwater reflections, CoReS gives the right mask. It is the chain-like hierarchy that results in correct and fine-grained segmentation, which cannot be achieved by LISA.

\subsection{Comparison on other benchmarks}
We also evaluate the performance of CoReS on the referring segmentation task, as shown in Tab.\ref{otherbenchmark}. As a degraded form of reasoning segmentation, the results of the referring segmentation are compared on refCOCO, refCOCO+, and refCOCOg, and its performance is measured using cIoU. For general models, we compare with X-Decoder and SEEM. For specialized models, we compare CoReS with methods like PolyFormer. Without using their REC pretraining and mass data from tasks like object detection and caption for targeted training, CoReS still achieves competitive results, as shown in Tab.\ref{otherbenchmark}. However, these methods could not handle the ReasonSeg tasks that CoReS is designed for.

\begin{table}[!t]
\caption{Performance comparison between CoReS and other methods on referring segmentation datasets. The LISA and CoReS here refer to their LLaVA-7B-v0 version. We use cIoU as the metric here. `S' and `G' refer to general/specialized types of models. `RS' refers to the ability to handle reasoning segmentation tasks.}
\label{otherbenchmark}
\centering
\setlength{\tabcolsep}{8pt}
\begin{threeparttable}
\resizebox{\linewidth}{!}{
\begin{tabular}{c|c|c|ccc|ccc|cc}
\toprule[1pt]
\multirow{2}{*}{Method}&\multirow{2}{*}{T}&\multirow{2}{*}{RS}&\multicolumn{3}{c|}{refCOCO}&\multicolumn{3}{c|}{refCOCO+}&\multicolumn{2}{c}{refCOCOg} \\
 & &&val & testA & testB& val & testA & testB& val(U) & testU \\
\midrule
MCN~\cite{luo2020multi}&S&&62.4&64.2&59.7&50.6&55.0&44.7&49.2&49.4\\
VLT~\cite{ding2022vlt}&S&&67.5&70.5&65.2&56.3&61.0&50.1&55.0&57.7\\
CRIS~\cite{wang2022cris}&S&&70.5&73.2&66.1&62.3&68.1&53.7&59.9&60.4\\
LAVT~\cite{yang2022lavt}&S&&72.7&75.8&68.8&62.1&68.4&55.1&61.2&62.1\\
ReLA~\cite{liu2023gres}&S&&73.8&76.5&70.2&66.0&71.0&57.7&65.0&66.0\\
PolyFormer~\cite{liu2023polyformer}&S&&74.8&76.6&71.1&\textbf{67.6}&\textbf{72.9}&\textbf{59.3}&67.8&69.1\\
X-Decoder~\cite{zou2023generalized}&G&&-&-&-&-&-&-&64.6&-\\
SEEM~\cite{zou2023segment}&G&&-&-&-&-&-&-&65.7&-\\
\midrule
LISA~\cite{lai2023lisa}&G&\checkmark&74.9&\textbf{79.1}&72.3&65.1&70.8&58.1&67.9&70.6\\
\rowcolor{gray!20}
CoReS&G&\checkmark&\textbf{76.0}&78.6&\textbf{72.5}&65.1&70.0&58.6&\textbf{69.0}&\textbf{70.7}\\
\bottomrule[1pt]     
\end{tabular}}
\end{threeparttable}
\end{table}

Compared to LISA, CoReS does not exhibit a significant improvement on refCOCO and refCOCO+, whose average query lengths are 3.6 and 3.5, respectively. However, on refCOCOg with an average query length of 8.4, our method outperforms LISA by about 2\%. We attribute this to the fact that CoReS's multi-modal chain of thought primarily enhances the performance under complex and challenging queries. Therefore, the improvement is less significant for simpler referential objects in the former two datasets, but more pronounced for datasets involving more complex object references such as refCOCOg. This demonstrates that the proposed multi-modal chain of thought can indeed reduce the difficulty of finding the referred instance by complex reasoning query, proving the effectiveness of CoReS. We include performance results on other benchmarks in the appendix.

\subsection{Ablation Studies}
\textbf{Different design choices of CoReS.}
To demonstrate the effectiveness of our proposed in-context input and the dual-chain-of-thought structure, we conduct ablation experiments on the model design details mentioned for the CoReS in Tab.\ref{ablation} and different design alternatives in the Appendix. 

In the case of using chain-of-segmenting without chain-of-reasoning, we use the same \texttt{[SEG]} embedding as the prompt input for each level, resulting in only a 1.1\% improvement in gIoU. This demonstrates that the improvement brought by our method is not due to an increase in computational power. When using chain-of-reasoning without the segmentation chain, to avoid information loss, we take the average of the two token embeddings as the input for the mask decoder, resulting in a 1.0\% boost, demonstrating that the decomposition of thinking levels can indeed reduce the difficulty and improve performance. 

\begin{table}[!t]
\caption{Ablation studies on the key components of CoReS. ``InC'' and ``CoR'' refer to the in-context input and the chain-of-reasoning. ``CoS'' and ``CoS-R'' denote the chain-of-segmenting without and with the token refinement, respectively.}
\label{ablation}
\centering
\setlength{\tabcolsep}{8pt}
\begin{threeparttable}
\begin{tabular}{cccc |c c}
\toprule[1pt]
\multicolumn{4}{c|}{Components} &\multicolumn{2}{c}{ReasonSeg} \\
InC&CoR&CoS&CoS-R & \makebox[0.1\textwidth][c]{gIoU} & \makebox[0.1\textwidth][c]{cIoU} \\
\midrule
&&&&52.9&54.0   \\
\checkmark&&&&53.2&55.4   \\
&\checkmark&&&53.9&55.6 \\
&&\checkmark&&54.0&54.1  \\
\checkmark&\checkmark&&&55.3&56.9    \\
\checkmark&&\checkmark&&55.4&58.0    \\
&\checkmark&\checkmark&&56.9&59.8    \\
&\checkmark&&\checkmark&57.5&60.4    \\
\checkmark&\checkmark&\checkmark&&58.4&59.3   \\
\rowcolor{gray!20}
\checkmark&\checkmark&&\checkmark&\textbf{59.4}&\textbf{62.1}   \\
\bottomrule[1pt]     
\end{tabular}
\end{threeparttable}
\end{table}

The CoS guided by CoR leads to a 4.6\% performance increase, proving the effective interaction and guidance between the two modalities through chain thinking. By adding in-context input to the dual-chain structure, CoReS can continue to gain a 1.9\% improvement. This demonstrates that the strategy of in-context training can provide prompts and guidance for the dual-chain structure. Consistent with Sec.\ref{sec:method}, this proves that our practices are reasonable and reliable.

\textbf{Different numbers of the in-context input.}
We conduct experiments with different numbers of the in-context input to further explore their effect. As shown in Tab.\ref{icl-num}, adding one in-context example can lead to a 1.9\% improvement. However, as the number of examples continues to increase, performance improvement is slightly decreased. This may be due to the limited quality of the context library, which may not be conducive to the performance. In general, regardless of the number of in-context examples added, compared to not using in-context input, it always leads to improved results. This proves the effectiveness of the logical guidance of in-context inputs. 

\textbf{Different logical levels.} We also look into the effect of different logical levels of the dual chain of thought. In Sec.\ref{sec:method} and Tab.\ref{mainexcel}-Tab.\ref{icl-num}, CoReS selects a logical depth of 2, resulting in a 5.8\% improvement in gIoU, where both the in-context input and the dual-modal chain adhere to a two-level semantic structure from \texttt{[LOC]} to \texttt{[SEG]}. 

\begin{minipage}{\textwidth}
\begin{minipage}[t]{0.45\textwidth}
\makeatletter\def\@captype{table}
\caption{Ablation studies of different numbers of the in-context input.}
\label{icl-num}
\centering
\setlength{\tabcolsep}{8pt}
\begin{threeparttable}
\begin{tabular}{c|cc }
\toprule[1pt]
\multirow{2}{*}{num}&\multicolumn{2}{c}{ReasonSeg}   \\
 & \makebox[0.2\textwidth][c]{gIoU} & \makebox[0.2\textwidth][c]{cIoU} \\
\midrule
0&57.5&60.4   \\
\rowcolor{gray!20}
1&\textbf{59.4}&\textbf{62.1}	   \\
2&59.4&61.9  \\
4&59.3&62.0  \\
\bottomrule[1pt]     
\end{tabular}
\end{threeparttable}
\end{minipage}
\begin{minipage}[t]{0.45\textwidth}
\makeatletter\def\@captype{table}
\caption{Ablation studies of different logical levels. [S], [P], and [L] refer to the \texttt{[SEG]}, \texttt{[POS]}, and \texttt{[LOS]} token.}
\label{chainlevel}
\centering
\setlength{\tabcolsep}{8pt}
\begin{threeparttable}
\begin{tabular}{c|cc }
\toprule[1pt]
\multirow{2}{*}{token}&\multicolumn{2}{c}{ReasonSeg}  \\
 & gIoU & cIoU  \\
\midrule[1pt]
{$[S]$}&53.2&55.4   \\
{$[L]+[S]$}&59.4&\textbf{62.1}	   \\
\rowcolor{gray!20}
{$[L]+[P]+[S]$}&\textbf{59.7}&60.8  \\
\bottomrule[1pt]     
\end{tabular}
\end{threeparttable}
\end{minipage}
\end{minipage}

Similarly, we test the performance of CoReS with a three-level semantic hierarchy. For the corresponding levels, we heuristically select ``commonly occurring categories'', ``specific locations in the image'' and ``intrinsic features''. The template of the chain-of-reasoning, the mask proposal process, and the in-context inputs are also adjusted accordingly to the three levels. The performance does not increase much. We deem the reason to be that questions in ReasonSeg are not so particularly complex that require three levels of thinking and grounding. Furthermore, the specific content of each level of the three-level logic is heuristically, and its scientificity and reasonableness need to be considered. Besides, the longer logic chains itself is associated with increased backpropagation difficulty and slower convergence speed. However, overall, multi-level logic does bring a significant improvement in performance compared to single-plane logic, which is all consistent with the theoretical analysis in Sec.\ref{sec:method}.
\section{Conclusion}\label{sec:conclusion}
We propose CoReS, a dual-modal chain-of-thought framework for fine-grained reasoning tasks. Adopting a top-down logic for visual search, we propose a dual-chain hierarchical structure that assists the MLLM in accurately localizing objects referred to in reasoning texts. Additionally, the integration of in-context inputs enables the MLLM to achieve rule transfer, guiding the generation of chain-like outputs in multi-modal tasks. Experimental results confirm that our approach yields a substantial improvement over state-of-the-art methods in reasoning segmentation tasks. We hope that CoReS will inspire future research on a broader range of complex multi-modal tasks.
\clearpage
\bibliographystyle{splncs04}
\bibliography{main.bib}

\clearpage
\appendix
\renewcommand\thesection{\Alph{section}}
\renewcommand\thefigure{S\arabic{figure}}
\renewcommand\thetable{S\arabic{table}}
\renewcommand\theequation{S\arabic{equation}}
\setcounter{figure}{0}
\setcounter{table}{0}
\setcounter{equation}{0}

\section*{Appendix}

\section{Implementation Details}\label{appendix:implementation}
\textbf{For training datasets,} we follow LISA to utilize a combination of datasets for semantic segmentation, COCO-Stuff, PACO-LVIS, PASCAL-Part, and Mapillary Vistas), referring segmentation(refCLEF, refCOCO, refCOCO+, and refCOCOg), and the ReasonSeg training set for fine-tuning. We abandon using the visual question answering (LLaVA-Instruct-150k) dataset employed by the baseline method LISA. The rationale lies in CoReS's control over the chain-of-reasoning process through the fixed templates of LLM output formats. The textual outputs from the VQA dataset, which may conflict with the logic of template-based outputs and fine-grained tasks, could adversely affect the training process.

\textbf{For the test dataset}, we conduct experiments on the ReasonSeg. It comprises image-question pairs with reasoning difficulty and the ground truth segmentation masks. Its training and validation set includes 239 and 200 image-instruction pairs, respectively.

As for the LoRA tuning of the MLLM, the LoRA rank is set to 8 and the LoRA alpha and dropout are set to 16 and 0.05, respectively. We apply deepspeed as the code engine, with 8 NVIDIA A100 for training. The optimizer for CoReS is AdamW and its learning rate is set to 0.0003, with a 100-step warm-up-decay. 

\section{Ablation Experiments}\label{appendix:benchmarks}
As for different design alternatives of CoReS, we conduct experiments in Tab.\ref{ablationCoT}. \textbf{Text-Only CoT}: We first compare with a CoT-version of LISA which performs explicit text-form reasoning before segmentation, which is trained with the additional text generated by GPT as supervision. This Text-Only CoT decreases the performance of LISA. One possible reason is that textual outputs weaken the info in [SEG], consequently impacting the subsequent segmentation process. \textbf{Supervised C1}, we choose two ways to provide supervision for C1, \textcircled{1}`\textit{C1-GT}' is using the final mask for C1 loss calculation at a low weight scale. The decrease in performance can be attributed to the undermining of the CoT formation, regardless of weight differences. \textcircled{2}`\textit{Pseudo}' refers to generating larger masks as C1 gt. It can be observed that leading MLLM to generate without constraints is better than rigid constraints. \textbf{Query-Specific Textual Prompts}: We adopt query-specific textual prompts (instead of irrelevant prompts) as inputs. Although it may not be practical in applications due to time and cost considerations, the improvement, as a rough upper bound, indirectly validates the effectiveness of in-context input.

\begin{table}[!h]
\caption{Implementation strategies of CoT. `LISACoT' refers to the text-only version of CoT, and `relevant' refers to using the trained CoReS-7B with query-related prompts generated by GPT. `C1-GT' and `pseudo' are the above two C1 supervisions.}
\setlength{\abovecaptionskip}{0cm}
\setlength{\belowcaptionskip}{-0.2cm}
\centering
\setlength{\tabcolsep}{8pt}
\begin{threeparttable}
\begin{tabular}{c|c|c|c|c|c|c}
\toprule[1pt]
Metric&CoReS&LISA&LISACoT&relevant&C1-GT&pseudo\\
\midrule
gIoU&59.4&52.9&44.7&60.1&55.9&58.1\\
cIoU&62.1&54.0&45.9&62.8&55.8&62.1\\
\bottomrule[1pt]     
\end{tabular}
\end{threeparttable}
\label{ablationCoT}
\end{table}

We conduct experiments on the hyperparameters in Tab.\ref{ablation2}. Placing excessive emphasis on the reasoning chain loss leads to a decrease in model performance. This is because it merely serves as an implicit constraint on the output of intermediate text, while the ultimate segmentation chain loss deserves more attention. Furthermore, increasing the weight of the DICE loss also has a positive effect on the results, which aligns with empirical intuition.

\begin{table}[!h]
\caption{Hyperparameters analysis of loss weights.}
\label{ablation2}
\centering
\setlength{\tabcolsep}{8pt}
\begin{threeparttable}
\begin{tabular}{l|ccc|ccc}
\toprule[1pt]
Hyper- & \multicolumn{3}{c|}{$\lambda_{COS}/\lambda_{COR}$}& \multicolumn{3}{c}{$\lambda_{CE}/\lambda_{DICE}$}\\
param&2:1&1:1&1:2&4:1&3:2&2:3\\
\midrule
gIoU&\textbf{59.4}&58.8&55.5&57.6&59.0&\textbf{59.4}\\
cIoU&\textbf{62.1}&61.1&58.3&62.2&\textbf{66.2}&62.1\\
\bottomrule[1pt]     
\end{tabular}
\end{threeparttable}
\end{table}

We use gIoU as our primary metric, rather than the cIoU with a bias towards large-area targets. To clarify, we divide target objects based on their sizes, as shown in Tab.B. cIoU is dominated by the performance on large objects, which dilutes the segmentation advantage of CoReS on more challenging small targets.

\begin{minipage}{\textwidth}
\begin{minipage}[t]{0.5\textwidth}
\makeatletter\def\@captype{table}
\caption{Ablation of different metrics. `$cIoU_s$', `$cIoU_m$' and `$cIoU_l$' refer to cIoU of objects with small, medium, and large sizes, respectively.}
\label{otherbenchmark}
\centering
\begin{threeparttable}
\begin{tabular}{c|ccccc}
\toprule[1pt]
\multirow{2}{*}{Method}&\multicolumn{5}{c}{LLaVA-v1.5-13B}\\
&gIoU&cIoU&$cIoU_s$&$cIoU_m$&$cIoU_l$ \\
\midrule
LISA&65.0&67.8&22.4&53.6&78.7\\
\rowcolor{gray!20}
CoReS&\textbf{68.1}&\textbf{68.2}&\textbf{35.7}&\textbf{58.5}&\textbf{80.5}\\
\bottomrule[1pt]     
\end{tabular}
\end{threeparttable}
\end{minipage}
\begin{minipage}[t]{0.45\textwidth}
\makeatletter\def\@captype{table}
\caption{Performance comparison on part segmentation datasets. `PIN', `PP', and `RP' are PartImageNet, PascalPart, and ReasonPart, respectively.}
\label{others}
\centering
\begin{threeparttable}
\begin{tabular}{c|cc|cc}
\toprule[1pt]
\multirow{2}{*}{Method}&PIN&PP&\multicolumn{2}{c}{RP} \\
&mIoU&mIoU&gIoU&cIoU\\
\midrule
LISA&33.3&14.8&14.1&18.9   \\
\rowcolor{gray!20}
CoReS&\textbf{39.0}&\textbf{19.6}&\textbf{20.9}&\textbf{33.1}  \\
\bottomrule[1pt]     
\end{tabular}
\end{threeparttable}
\end{minipage}
\vspace{6pt}
\end{minipage}

\section{Experiments on other benchmarks}\label{appendix:benchmarks}

Additionally, we evaluate the ability of CoReS on general fine-grained segmentation datasets. The mean-intersection-over-union (mIoU) serves as the evaluation metric here. On PartImageNet, the zero-shot performance of CoReS shows a 5.7\% improvement over LISA. Similarly, on the validation set of PascalPart, an approximate 5\% enhancement is observed. These results demonstrate that the multimodal reasoning chain structure of CoReS not only elevates the understanding of reasoning tasks but also augments the capability to comprehend details in general dense visual tasks.

In addition, we construct two reasoning-based part segmentation benchmarks based on the aforementioned two component-level datasets. We employ in-context learning with chatGPT to generate questions about category names, yielding questions such as "In the oceanic scene, which functional feature aids fish in steering and maintaining stability in the currents?" ReasonPart is established on 2,957 validation set images from PartImageNet and 4,465 images from PascalPart. As Tab.\ref{others} indicates, on the challenging benchmark ReasonPart, which features complex queries and fine-grained segmentation parts, CoReS outperforms LISA by 6.8\% in terms of gIoU and approximately 15\% in cIoU, substantiating the efficacy of its dual-chain structure.

\section{Analysis of C1 masks}\label{appendix:Quantitative}
Obtaining mask annotations for intermediate logical layers is impractical due to the logic subjectivity, so MLLM with world knowledge is used for a unified unsupervised representation. The scene-level info-injection of C1 is firstly based on MLLM's in-context learning ability, where in-context inputs are provided for logical prompts. Secondly, the MLLM's semantically coherent and reasonable output helps the formation of the hierarchy. Supervised templates, like `It appears on [LOC],' constrain the output semantics after `appears on' naturally carry the scene-level or item location meaning. Thirdly, since a rough mask is required as SAM input, the gradient backpropagation also pushes [LOC] toward the scene-level info.

\section{Quantitative Results}\label{appendix:Quantitative}

\begin{figure}[!t]
  \centering
  \includegraphics[width=\linewidth]{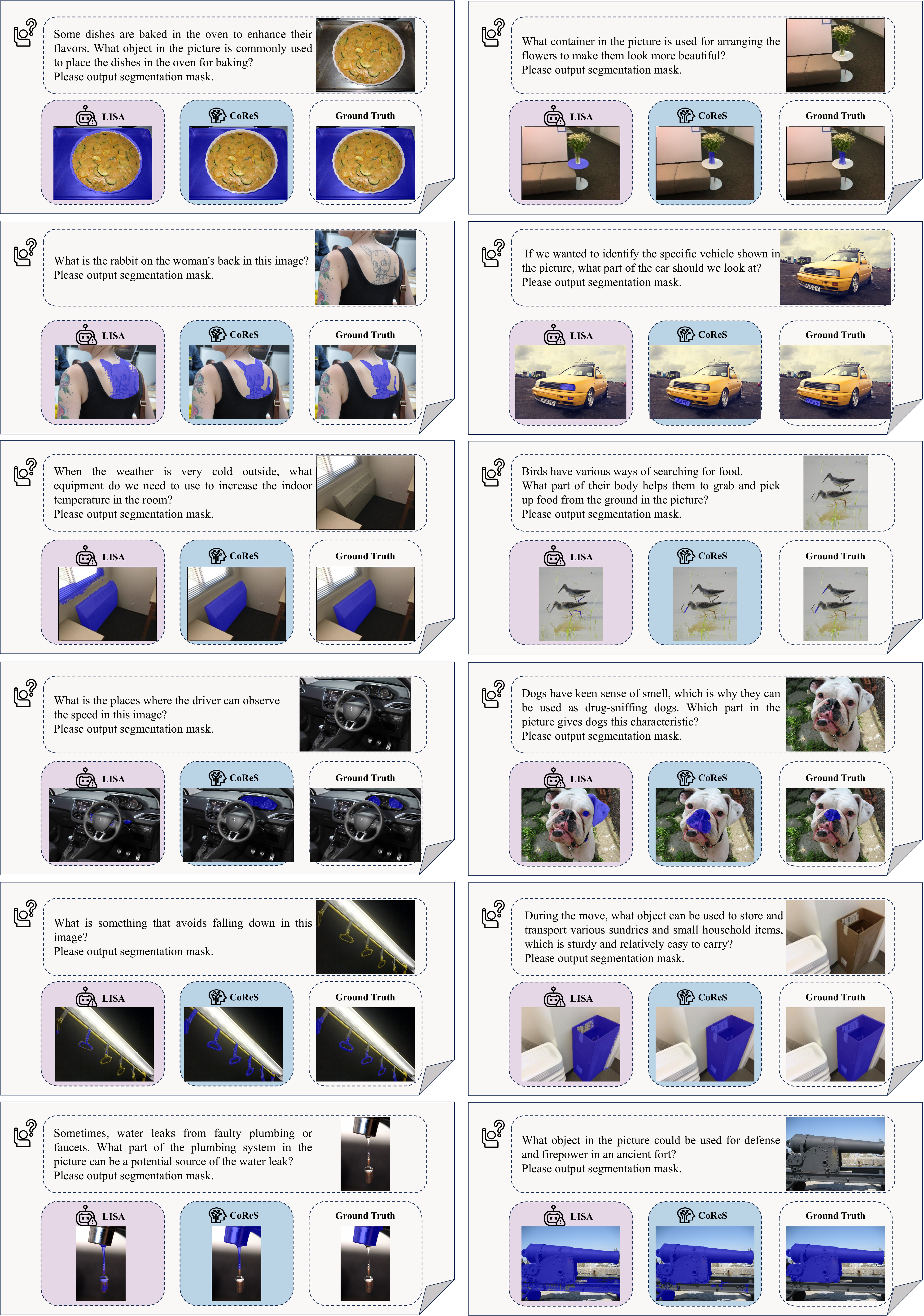}
  \caption{Visual comparison of CoReS and LISA.}
  \label{append2}
  \vspace{5pt}
\end{figure}

\begin{figure}[!t]
  \centering
  \vspace{5pt}
  \includegraphics[width=\linewidth]{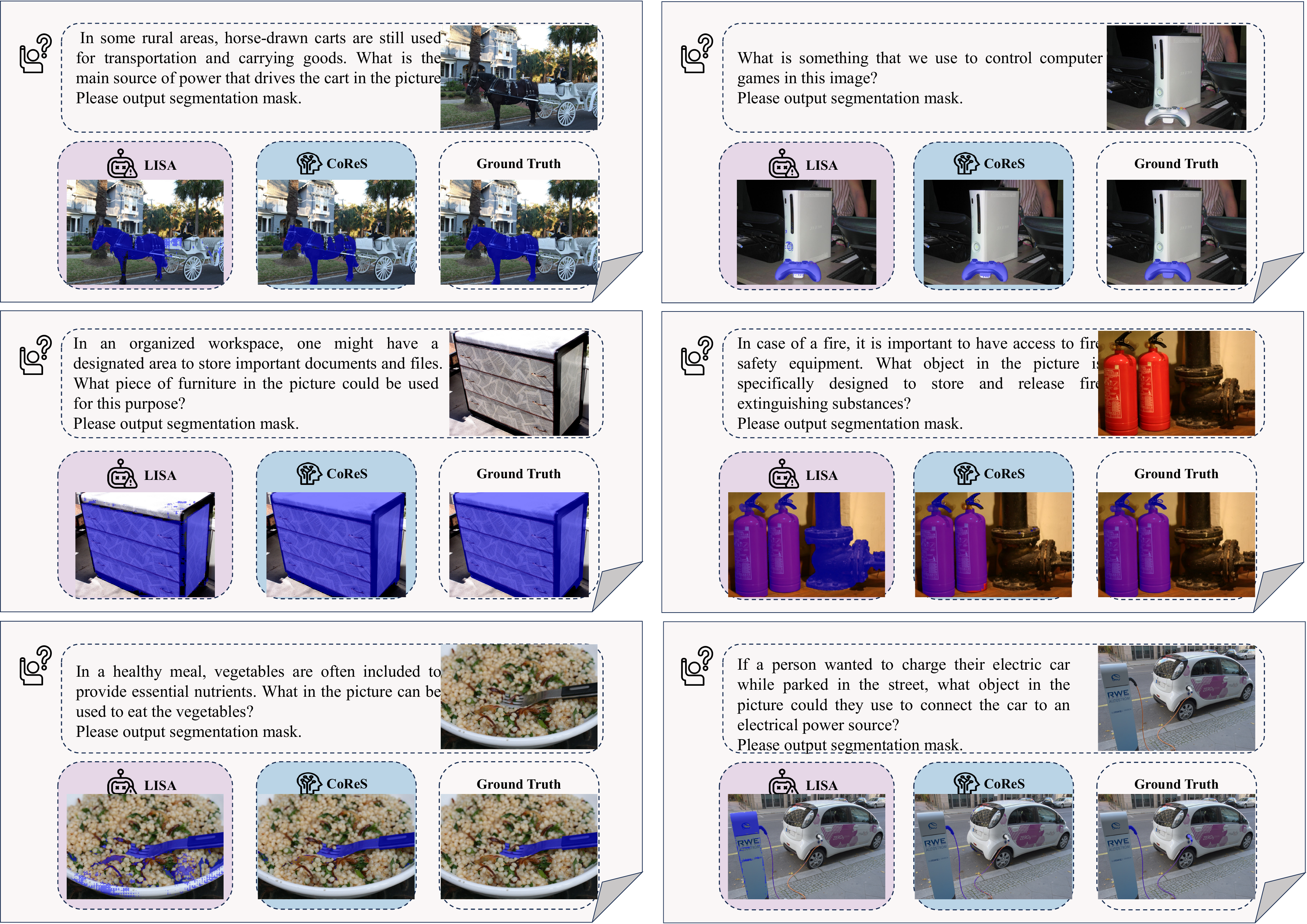}
  \caption{Visual comparison of CoReS and LISA.}
  \label{append1}
  \vspace{5pt}
\end{figure}

Qualitative results also prove the effectiveness of CoReS in Fig.\ref{append1} and Fig.\ref{append2}. When facing complex reasoning problems, LISA exhibits errors in grounding the exact instance referred by the reasoning query, while CoReS achieves the correct answers. Whether it's objects that are difficult to segment, such as a fork inserted into rice, or challenging reasoning tasks like determining the source of power for carts in an image, the chain-like hierarchy of CoReS results in correct and fine-grained segmentation.

\end{document}